\documentclass[conference, a4paper]{IEEEtran}
\IEEEoverridecommandlockouts
\usepackage[top=0.70in,bottom=1in,left=0.62in,right=0.62in]{geometry}
\usepackage{cite}
\usepackage{amsmath,amssymb,amsfonts}
\usepackage{graphicx}
\usepackage{fix-cm}
\usepackage{subfigure}
\usepackage{textcomp}
\usepackage{xcolor}
\usepackage{eso-pic}
\usepackage{url}
\usepackage{multicol}
\usepackage{multirow}
\usepackage{amsmath}
\usepackage{graphicx}
\usepackage{lipsum}
\usepackage{fancybox}
\usepackage{listings}
\usepackage[utf8]{inputenc}
\usepackage{comment}
\usepackage{hyperref}
\usepackage{url}
\usepackage{adjustbox}
\usepackage{array}
\usepackage{fancyhdr}
\def\BibTeX{{\rm B\kern-.05em{\sc i\kern-.025em b}\kern-.08em
    T\kern-.1667em\lower.7ex\hbox{E}\kern-.125emX}}

\makeatletter

\def\ps@IEEEtitlepagestyle{%
    \def\@oddfoot{\mycopyrightnotice}%
    \def\@evenfoot{}%
}
\def\mycopyrightnotice{
  {\footnotesize 979-8-3315-4791-2/26/\$31.00 ~\copyright2026 ~IEEE\hfill} 
  \gdef\mycopyrightnotice{}
}

\let\old@ps@IEEEtitlepagestyle\ps@IEEEtitlepagestyle
\def\confheader#1{%
    \def\ps@IEEEtitlepagestyle{%
        \old@ps@IEEEtitlepagestyle%
        \def\@oddhead{\strut #1\hfill}%
        \def\@evenhead{\strut #1\hfill}%
    }%
    \ps@headings%

}
\makeatother

\begin{document}
\newcommand\AtPageUpperMyright[1]{\AtPageUpperLeft{
 \put(\LenToUnit{0.07\paperwidth},\LenToUnit{-1cm}){ 
     \parbox{0.78\textwidth}{\raggedright\fontsize{9}{11}\selectfont #1}} 
 }}


\title{Explainable Machine Learning Framework for Cardiovascular Disease Diagnosis and Prognosis \\

}
\author{\IEEEauthorblockN{\textsuperscript{} Md. Emon Akter Sourov}
\IEEEauthorblockA{\textit{Dept. of Computer Science and}\\ Engineering \\
\textit{Bangladesh University,}\\
Dhaka-1207, Bangladesh. \\
emonsourov@gmail.com}
\and
\IEEEauthorblockN{\textsuperscript{} Md. Sabbir Hossen}
\IEEEauthorblockA{\textit{Dept. of Computer Science and}\\ Engineering \\
\textit{Bangladesh University,}\\
Dhaka-1207, Bangladesh. \\
sabbirhossen5622@gmail.com}
\and
\IEEEauthorblockN{\textsuperscript{} Pabon Shaha}
\IEEEauthorblockA{\textit{Dept. of Computer Science and}\\ Engineering \\
\textit{Mawlana Bhashani Science and Technology}\\ University,
Santosh, Tangail-1902. \\
pabonshahacse15@gmail.com}
\and
\IEEEauthorblockN{\textsuperscript{} Md. Moradul Siddique}
\IEEEauthorblockA{\textit{Dept. of Computer Science and Engineering} \\
\textit{University of Information Technology }\\and Sciences,
 Dhaka, Bangladesh. \\
moradul\_siddique@uits.edu.bd}
\and
\IEEEauthorblockN{\textsuperscript{} Yadab Sutradhar}
\IEEEauthorblockA{\textit{Dept. of Computer Science} \\
\textit{Maharishi International University, }\\
Fairfield, IA 52557, United States. \\
yadab.sutradhar@miu.edu}
\and
\IEEEauthorblockN{\textsuperscript{} Md Sadiq Iqbal}
\IEEEauthorblockA{\textit{Dept. of Computer Science and}\\Engineering \\
\textit{Bangladesh University,}\\
Dhaka-1207, Bangladesh. \\
sadiq.iqbal@bu.edu.bd}
}

\maketitle

\begin{abstract}
Heart disease continues to pose a critical worldwide health issue, more specifically in areas with insufficient access to healthcare infrastructure and diagnostic systems. Conventional diagnostic approaches often fall short in accurately detecting and managing heart disease risks, resulting in unfavorable outcomes. Machine learning presents a powerful means to boost the precision and reliability of cardiovascular disease prognosis and diagnosis. In this research, we introduced a unified approach incorporating classification techniques for detecting heart disease and regression techniques for forecasting associated risks. The analysis utilized the dataset, named Heart Disease, containing 1,035 instances. To mitigate the problem of data disproportion, the SMOTE was implemented, producing 100,000 additional synthetic samples. Evaluation metrics such as F1-score, recall, precision, accuracy, MAE, RMSE, MSE, and R² were adopted to evaluate the performance of the models. Among the classification algorithms, Random Forest delivered the most notable results, attaining an accuracy of 0.972 on actual data and 0.976 on artificially generated data. For prediction modeling, for both synthetic and real samples, linear regression produced the best R2 values of 0.992 and 0.984, respectively, along with the least amount of measurement errors. Furthermore, Explainable AI methods were utilized to improve the comprehensibility of the model outcomes. This paper emphasizes the transformative capabilities of machine learning for diagnosing cardiovascular disease and estimating risk levels, thereby supporting timely interventions and enhancing clinical settings. \vspace{3mm}
\end{abstract}

\begin{IEEEkeywords}
disease detection, heart diseases, cardiovascular diseases, explainable ai, and machine learning. \vspace{3mm} 
\end{IEEEkeywords}

\section{INTRODUCTION}
The global incidence of heart disease continues to have a startlingly high prevalence. The phrase “Heart Disease Risk” refers to the likelihood of individuals facing complications or negative health outcomes from cardiovascular issues, including coronary artery disease, heart failure, or arrhythmias. These hazards not only endanger personal health and lives but also place immense strain on healthcare infrastructures and national economies \cite{chandola2008work}. Heart illnesses are progressively prevalent in low-resource environments, where diagnostic capabilities and specialist treatments are often unavailable \cite{hoffman2013global}. As noted by the World Health Organization (WHO) in 2023 \cite{tribune}, Heart disease is a significant public health hazard in Bangladesh, resulting in 273,000 deaths annually. Of these, cardiac attack is the leading cause, responsible for 34\% of total national mortality. Risk factors for cardiovascular illness appear in multiple forms, including high cholesterol, diabetes, obesity, hypertension, and lifestyles like unhealthy diets and smoking. These are frequently intensified by deeper contributors like hereditary traits, insufficient preventive care, and limited awareness or delayed intervention \cite{mondesir2016diabetes}. Since these dangers frequently develop covertly, early identification and consistent monitoring are essential \cite{mocumbi2011challenges}. Machine learning has become a potent tool in recent years for identifying a variety of illnesses across multiple domains, such as healthcare and agriculture \cite{hossen2025brain, strawberry}. It has shown potential in forecasting cardiovascular risk and identifying heart disease in early phases. This provides a novel method for improving cardiac healthcare outcomes \cite{chang2022artificial}. 
Major Contributions of this study-


\begin{itemize}
    \item An explainable machine learning framework combining classification and regression models to evaluate heart disease risk, offering an integrated solution for immediate diagnosis and prognosis.
    \item A set of ten classification  model like TabNet, SVM, DT, RF, KNN, NB, GB, XGB, LXGB, and catBoost  along with eleven regression models, including sophisticated models such as Lasso,  Ridge, TabNet, LightGBM, and CatBoost regressor was utilized.
    \item The Synthetic Minority Oversampling Technique (SMOTE) was used to rectify the imbalance in the data.
    \item The ML algorithms underwent a thorough evaluation using performance indicators like F1-score, recall, precision, accuracy, MCC, R², MSE, RMSE, and MAE for comprehensive assessment. Furthermore, Explainable AI methods were introduced to make the models’ forecasts more interpretable and trustworthy for clinical use. 
\end{itemize}

The rest of the paper is organized as follows: Section II reviews related work and highlights existing limitations. Section III describes the dataset and the proposed methodology for cardiovascular disease detection and risk analysis using machine learning. Section IV presents and discusses the experimental results, while Section V concludes the study and outlines future research directions.
\vspace{4mm}

\section{LITERATURE REVIEW}
\vspace{1mm}
Many researchers have proposed machine learning models to predict cardiovascular disease risk, highlighting its importance for improving clinical outcomes. Despite significant progress, it remains an active and evolving research area. Several previous studies were reviewed to understand current developments and challenges in heart disease diagnosis and prediction. 
\textit{Rabbi et al.} \cite{pp1} constructed an ensemble-based approach combining GNB, DT, LR, KNN, SVM, and RF with advanced techniques like stacking, bagging, voting, and boosting. Evaluated on the Cleveland, Indicators of Heart Disease, and Framingham datasets, the bagging ensemble achieved a top accuracy of 97\% on Framingham and Indicators, while the voting ensemble reached 92\% on Cleveland. The proposed BEMLA consistently outperformed individual classifiers, offering a robust solution for heart disease prediction. \textit{Ganie et al.} \cite{pp2} enhanced cardiovascular disease forecasting using voting and stacking ensembles derived from 15 base algorithms trained on a couple of datasets. Six optimal models were combined into meta-ensembles, with stacking achieving the best performance. Statistical tests, including Friedman and Holm’s post-hoc, validated the models’ superiority. SHAP-based XAI was employed for interpretability, showing how feature contributions impact predictions. \textit{Rohan et al.} \cite{pp3} conducted an extensive evaluation involving 11 feature selection techniques and 21 classifiers for heart disease prediction. Models included CNN, LSTM, GRU, BiLSTM, RF, SVM, XGBoost, and more. XGBoost attained the most extraordinary performance, with 97\% accuracy, 98\% sensitivity, and an F1-score of 0.98, outperforming all others across multiple metrics. \textit{Nissa et al.} \cite{r16} emphasized fast classification using boosting models like AdaBoost, LightGBM, Gradient Boosting, RF, and CatBoost. AdaBoost achieved the top performance in their study with 95\% accuracy, though tuning and evaluation issues were acknowledged that might impact performance generalizability. \textit{Singh et al.} \cite{r10} applied machine learning techniques to predict congestive heart failure using a reduced set of features to lower diagnostic costs and improve accuracy. They combined KNN and the C4.5 algorithm for feature optimization and handling missing data. The study compared DNN with six ML models (RF, SVM, NB, DT, LR, and KNN), where DNN achieved the best performance with 95.30\% accuracy.  \textit{Husnain et al.} \cite{r12} showcased the potential of artificial intelligence to forecast heart diseases using methods like neural networks SVM, and RF. The neural network surpassed conventional diagnostic tools, achieving 92\% accuracy for high-risk patient identification. \textit{Mienye et al.} \cite{r1} introduced a method that integrates SHAP-based interpretability, Bayesian hyperparameter optimization, and robust ensemble techniques. Ensemble models, including AdaBoost, RF, and XGBoost, were evaluated. Their optimized XGBoost achieved notable results on the Cleveland dataset, with 0.971 specificity and 0.989 sensitivity. Nonetheless, the study’s strong reliance on Bayesian optimization may not ensure peak results across all data. \textit{Abuhaija et al.} \cite{r15} explored seven classifiers, multilayer perceptron, artificial neural network, LR, SVM, DT, KNN, and Naïve Bayes. A correlation-based filter was used to determine key features. Their evaluation based on precision, accuracy, specificity, and sensitivity showed that the J48 decision tree attained the greatest accuracy of 95.76\%. \textit{Bhatt et al.} \cite{r14} developed a predictive method to lower heart diseases mortality. They used Huang initialization with k-modes clustering and tested models such as RF, DT, XGBoost, and Multilayer Perceptron on a Kaggle dataset of 70,000 entries. The cross-validated MLP model achieved the highest accuracy of 87.28\%, outperforming others. 
\textit{Chandrasekhar et al.} \cite{chandrasekhar2023enhancing} utilized six ML models for heart disease forecasting using the IEEE Dataport and Cleveland datasets. AdaBoost attained 90\% accuracy on the IEEEDataport, whereas LR attained 90.16\% accuracy on the Cleveland. A soft voting ensemble increases the accuracy to 95\% and 93.44\%, respectively.

\section{MATERIALS \& METHODS}
\begin{figure} [htbp]
	\centering 
	\fbox{\includegraphics[height=4.5cm, width=8.4cm]{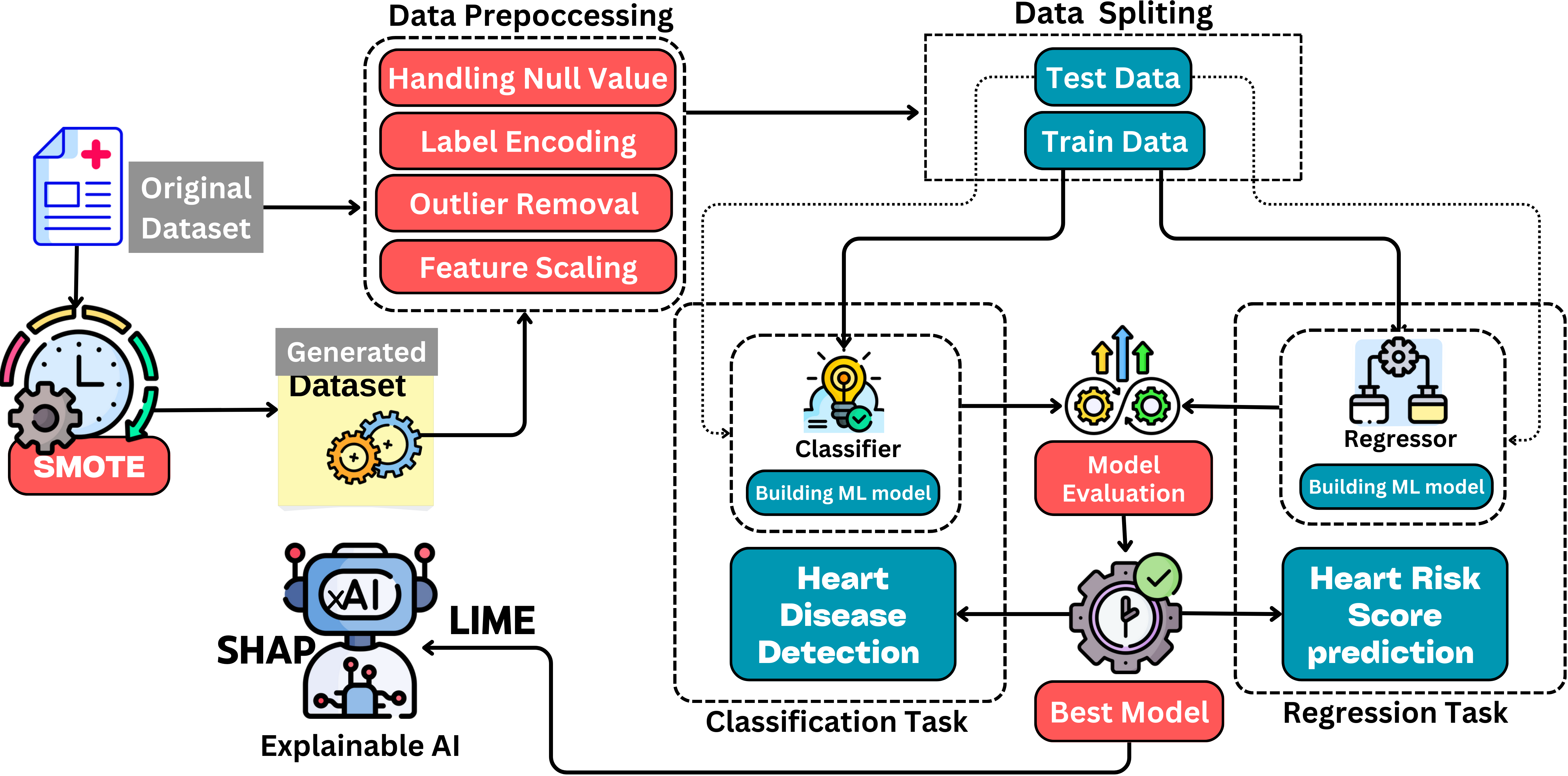}}	
		\caption{Overall Methodology Diagram for Machine Learning-Based Cardiovascular Disease Diagnosis and Prognosis} 
	\label{fig:md}
\end{figure}

\vspace{-2mm}
This study focuses on detecting heart disease and predicting related risks through both classification and regression methods using the heart disease dataset. The methods used in this study are depicted in Figure \ref{fig:md}. The overall methodology covers model training, data preprocessing, and the entire structure of the suggested approach.
\vspace{-2mm}

\subsection{Dataset}
The “Heart Disease dataset” utilized in this research was sourced from “Kaggle” \cite{dataset}. It comprises patient-related information that is essential for evaluating and forecasting the probability of cardiovascular disease. The dataset includes 16 attributes, offering insights such as peak heart rate, cholesterol, blood pressure, sex, age, ECG readings, chest pain occurrence, and other vital features instrumental in assessing cardiac risk. It consists of 1,035 entries in total,  504 representing healthy cases, and 531 indicating cardiovascular illness. The dataset is complete without any null values, and every feature is provided as numerical values.

\subsection{Data Preprocessing}
We conducted several preprocessing steps to prepare the dataset effectively. To ensure interoperability with machine learning models, label encoding was used to transform categorical data into numerical form. Although the dataset did not contain any missing values, we defined an imputation strategy for potential gaps: median values for numerical fields and mode values for categorical fields. The Interquartile Range (IQR) approach was used to identify and eliminate outliers in continuous characteristics. Additionally, feature scaling was applied to normalize the dataset, ensuring that all attributes contributed proportionately during model training and no single feature dominated due to its scale.

\subsection{Synthetic Data Generation}
To mitigate the imbalance in class distribution, we applied the Synthetic Minority Over-sampling Technique (SMOTE). This method synthesizes new examples by interpolating between samples from the minority class and their nearest neighbors. Using SMOTE, we created an additional 100,000 synthetic instances to balance the dataset better. Subsequently, we divided both the original and artificially balanced datasets into testing and training sets using a 20:80 ratio.

\subsection{Model Training}
This section outlines the training procedure for both classification and regression models. First, the dataset was preprocessed, and SMOTE was applied to handle class imbalance. The experiments were then carried out in two stages classification and regression each evaluated before and after applying SMOTE. In the classification stage, ten machine learning models, including LightGBM, XGBoost, GB, KNN, TabNet, CatBoost, DT, RF, Gaussian NB, and SVM, were used to identify heart disease classes. In the regression stage, eleven models such as CatBoost, LightGBM, XGBR, KNNR, DTR, RFR, GBR, Lasso, Ridge, and SVR were employed to predict continuous risk values.
 \vspace{2mm}



 
\subsection{Performance Parameters}
To assess the suggested approach, a comprehensive set of evaluation metrics was employed. Classifiers were measured using balanced accuracy (Acc), accuracy, precision, recall, and F1-score. For regression models, the evaluation relied on metrics such as Matthews Correlation Coefficient (MCC), R-squared (R²), Mean Absolute Error (MAE), Root Mean Squared Error (RMSE), and Mean Squared Error (MSE), capturing both prediction quality and reliability. The definitions and mathematical expressions for each metric are detailed below. \vspace{2mm}

\noindent Accuracy is the percentage of accurately identified outcomes relative to the total predictions, serving as a general metric for classification performance.
\begingroup\scriptsize
\begin{equation}
\text{Accuracy} = \frac{TP + TN}{TP + TN + FP + FN}
\end{equation}
\endgroup

\noindent Precision reflects the classification model's effectiveness in detecting each class by computing the fraction of true positives among predicted positives. \vspace{2mm}
\begingroup\scriptsize
\begin{equation}
\text{Precision} = \frac{TP}{TP + FP}
\end{equation}
\endgroup

\noindent Recall, expressed as the ratio of true positives to the sum of false negatives and true positives, assesses the model's capacity to detect positive results.
\begingroup\scriptsize
\begin{equation}
\text{Recall} = \frac{TP}{TP + FN}
\end{equation}
\endgroup

\noindent The F1 Score is a statistic that calculates the harmonic mean of recall and precision to assess the classifier's overall performance.
\begingroup\scriptsize
\begin{equation}
\text{F1 Score} = 2 \times \frac{\text{Recall} \times \text{Precision}}{\text{Recall} + \text{Precision}}
\end{equation}
\endgroup

\noindent The Matthews Correlation Coefficient, or MCC, measures the accuracy of binary classifications by taking into account every component of the confusion matrix to produce a fair measurement even when datasets are unbalanced.
\begingroup\scriptsize
\begin{equation}
\text{MCC} = \frac{(TP \times TN) - (FP \times FN)}{\sqrt{(TP + FP)(TP + FN)(TN + FP)(TN + FN)}}
\end{equation}
\endgroup

\noindent The mean squared error (MSE), which represents the total prediction error, is calculated as the average of the squared discrepancies between actual and projected values.
\begingroup\scriptsize
\begin{equation}
\text{MSE} = \frac{1}{n} \sum_{i=1}^{n} (y_i - \hat{y}_i)^2
\end{equation}
\endgroup

\noindent The coefficient of determination, or R2 score, quantifies the percentage of the target variable's variance that the model can account for; values nearer 1 denote better performance.
\begingroup\scriptsize
\begin{equation}
R^2 = 1 - \frac{\sum_{i=1}^{n} (y_i - \hat{y}_i)^2}{\sum_{i=1}^{n} (y_i - \bar{y})^2}
\end{equation}
\endgroup

\noindent The Root Mean Squared Error (RMSE), which is the square root of MSE, evaluates the distribution of residuals to provide an interpretable indicator of prediction accuracy.
\begingroup\scriptsize
\begin{equation}
\text{RMSE} = \sqrt{\frac{1}{n} \sum_{i=1}^{n} (y_i - \hat{y}_i)^2}
\end{equation}
\endgroup

\noindent Mean Absolute Error (MAE) provides information about the average error of the model by calculating the mean absolute difference between actual and forecasted data.
\begingroup\scriptsize
\begin{equation}
\text{MAE} = \frac{1}{n} \sum_{i=1}^{n} |y_i - \hat{y}_i|
\end{equation}
\endgroup

\subsection{Explainable AI}
We used Explainable AI (XAI) methods, such as SHAP and LIME, to make the machine learning classification and regression models more straightforward to understand. LIME produced localized interpretations by perturbing input attributes and observing their influence on predictions. In contrast, SHAP delivered global insights by assigning contribution scores to each feature according to how it affects the model’s outputs \cite{dwivedi2023explainable , hossen2025july}. These tools helped clarify how various attributes shaped the model’s behavior, ensuring interpretability and offering clarity for domain experts in the decision-making process.

\begin{table*}[htbp]
\centering
\setlength{\tabcolsep}{7.1pt}
\renewcommand{\arraystretch}{1.1}
\caption{Classification Models Performance for Diagnosis of Cardiovascular Diseases Before and After Utilizing SMOTE}
\label{tab:cn}
\begin{tabular}{|c|cc|cc|cc|cc|cc|}
\hline
\multirow{2}{*}{\textbf{Classifiers}} & \multicolumn{2}{c|}{\textbf{F1-Score}} & \multicolumn{2}{c|}{\textbf{Recall}} & \multicolumn{2}{c|}{\textbf{Precision}} & \multicolumn{2}{c|}{\textbf{MCC}} & \multicolumn{2}{c|}{\textbf{Accuracy}} \\ \cline{2-11} 
                                & \multicolumn{1}{c|}{\textbf{Original}}  & \textbf{SMOTE} & \multicolumn{1}{c|}{\textbf{Original}}  & \textbf{SMOTE} & \multicolumn{1}{c|}{\textbf{Original}}  & \textbf{SMOTE} & \multicolumn{1}{c|}{\textbf{Original}}  & \textbf{SMOTE} & \multicolumn{1}{c|}{\textbf{Original}}  & \textbf{SMOTE} \\ \hline
TabNet                          & \multicolumn{1}{c|}{0.461} & 0.474 & \multicolumn{1}{c|}{0.484} & 0.521 & \multicolumn{1}{c|}{0.441} & 0.437 & \multicolumn{1}{c|}{-0.204} & -0.182 & \multicolumn{1}{c|}{0.404} & 0.418 \\ \hline
Naive Bayes                     & \multicolumn{1}{c|}{0.825} & 0.831 & \multicolumn{1}{c|}{0.842} & 0.858 & \multicolumn{1}{c|}{0.812} & 0.806 & \multicolumn{1}{c|}{0.628} & 0.642 & \multicolumn{1}{c|}{0.814} & 0.822 \\ \hline
CatBoost                        & \multicolumn{1}{c|}{0.886} & 0.883 & \multicolumn{1}{c|}{0.905} & 0.904 & \multicolumn{1}{c|}{0.869} & 0.865 & \multicolumn{1}{c|}{0.754} & 0.760 & \multicolumn{1}{c|}{0.876} & 0.880 \\ \hline
LightGBM                        & \multicolumn{1}{c|}{0.903} & 0.908 & \multicolumn{1}{c|}{0.932} & 0.925 & \multicolumn{1}{c|}{0.877} & 0.894 & \multicolumn{1}{c|}{0.792} & 0.812 & \multicolumn{1}{c|}{0.896} & 0.906 \\ \hline
XGBoost                         & \multicolumn{1}{c|}{0.917} & 0.914 & \multicolumn{1}{c|}{0.942} & 0.928 & \multicolumn{1}{c|}{0.896} & 0.902 & \multicolumn{1}{c|}{0.824} & 0.824 & \multicolumn{1}{c|}{0.910} & 0.910 \\ \hline
Gradient Boosting               & \multicolumn{1}{c|}{0.960} & 0.972 & \multicolumn{1}{c|}{0.966} & 0.976 & \multicolumn{1}{c|}{0.955} & 0.969 & \multicolumn{1}{c|}{0.916} & 0.942 & \multicolumn{1}{c|}{0.956} & 0.970 \\ \hline
KNN                             & \multicolumn{1}{c|}{0.871} & 0.884 & \multicolumn{1}{c|}{0.881} & 0.896 & \multicolumn{1}{c|}{0.863} & 0.874 & \multicolumn{1}{c|}{0.726} & 0.760 & \multicolumn{1}{c|}{0.862} & 0.878 \\ \hline
Decision Tree                   & \multicolumn{1}{c|}{0.970} & 0.972 & \multicolumn{1}{c|}{0.968} & 0.983 & \multicolumn{1}{c|}{0.971} & 0.965 & \multicolumn{1}{c|}{0.934} & 0.948 & \multicolumn{1}{c|}{0.968} & 0.974 \\ \hline
\textbf{Random Forest}          & \multicolumn{1}{c|}{\textbf{0.973}} & \textbf{0.977} & \multicolumn{1}{c|}{\textbf{0.964}} & \textbf{0.981} & \multicolumn{1}{c|}{\textbf{0.983}} & \textbf{0.974} & \multicolumn{1}{c|}{\textbf{0.944}} & \textbf{0.952} & \multicolumn{1}{c|}{\textbf{0.972}} & \textbf{0.976} \\ \hline
SVM                             & \multicolumn{1}{c|}{0.904} & 0.897 & \multicolumn{1}{c|}{0.915} & 0.911 & \multicolumn{1}{c|}{0.896} & 0.887 & \multicolumn{1}{c|}{0.796} & 0.790 & \multicolumn{1}{c|}{0.894} & 0.896 \\ \hline
\end{tabular}
\end{table*}

\begin{table*}[htbp]
\centering
\setlength{\tabcolsep}{11.7pt}
\renewcommand{\arraystretch}{1.1}
\caption{Regression Models Performance for Predicting Cardiovascular Diseases Before and After Utilizing SMOTE}
\label{tab:rg}
\begin{tabular}{|c|cc|cc|cc|cc|}
\hline
\multirow{2}{*}{\textbf{Regressors}} & \multicolumn{2}{c|}{\textbf{MAE}}                        & \multicolumn{2}{c|}{\textbf{RMSE}}                       & \multicolumn{2}{c|}{\textbf{MSE}}                        & \multicolumn{2}{c|}{\textbf{R²}}                         \\ \cline{2-9} 
                                & \multicolumn{1}{c|}{\textbf{Original}}  & \textbf{SMOTE} & \multicolumn{1}{c|}{\textbf{Original}}  & \textbf{SMOTE} & \multicolumn{1}{c|}{\textbf{Original}}  & \textbf{SMOTE} & \multicolumn{1}{c|}{\textbf{Original}}  & \textbf{SMOTE} \\ \hline
Lasso                           & \multicolumn{1}{c|}{1.382}          & 1.400              & \multicolumn{1}{c|}{1.670}          & 1.704              & \multicolumn{1}{c|}{2.793}          & 2.907              & \multicolumn{1}{c|}{0.388}          & 0.354              \\ \hline
Ridge                           & \multicolumn{1}{c|}{0.062}          & 0.108              & \multicolumn{1}{c|}{0.262}          & 0.314              & \multicolumn{1}{c|}{0.082}          & 0.118              & \multicolumn{1}{c|}{0.982}          & 0.974              \\ \hline
\textbf{Linear Regression}      & \multicolumn{1}{c|}{\textbf{0.036}} & \textbf{0.066}     & \multicolumn{1}{c|}{\textbf{0.184}} & \textbf{0.238}     & \multicolumn{1}{c|}{\textbf{0.034}} & \textbf{0.061}     & \multicolumn{1}{c|}{\textbf{0.984}} & \textbf{0.992}     \\ \hline
CatBoost                        & \multicolumn{1}{c|}{0.858}          & 0.868              & \multicolumn{1}{c|}{1.068}          & 1.078              & \multicolumn{1}{c|}{1.144}          & 1.167              & \multicolumn{1}{c|}{0.750}          & 0.982              \\ \hline
LightGBM                        & \multicolumn{1}{c|}{0.076}          & 0.098              & \multicolumn{1}{c|}{0.278}          & 0.262              & \multicolumn{1}{c|}{0.087}          & 0.081              & \multicolumn{1}{c|}{0.978}          & 0.982              \\ \hline
XGBoost                         & \multicolumn{1}{c|}{0.030}          & 0.054              & \multicolumn{1}{c|}{0.264}          & 0.254              & \multicolumn{1}{c|}{0.074}          & 0.073              & \multicolumn{1}{c|}{0.984}          & 0.984              \\ \hline
Gradient Boosting               & \multicolumn{1}{c|}{0.120}          & 0.150              & \multicolumn{1}{c|}{0.292}          & 0.308              & \multicolumn{1}{c|}{0.091}          & 0.104              & \multicolumn{1}{c|}{0.978}          & 0.976              \\ \hline
KNN                             & \multicolumn{1}{c|}{0.484}          & 0.436              & \multicolumn{1}{c|}{0.706}          & 0.656              & \multicolumn{1}{c|}{0.504}          & 0.440              & \multicolumn{1}{c|}{0.888}          & 0.902              \\ \hline
Decision Tree                   & \multicolumn{1}{c|}{0.030}          & 0.060              & \multicolumn{1}{c|}{0.256}          & 0.274              & \multicolumn{1}{c|}{0.069}          & 0.087              & \multicolumn{1}{c|}{0.986}          & 0.980              \\ \hline
Random Forest                   & \multicolumn{1}{c|}{0.084}          & 0.102              & \multicolumn{1}{c|}{0.278}          & 0.278              & \multicolumn{1}{c|}{0.083}          & 0.086              & \multicolumn{1}{c|}{0.980}          & 0.980              \\ \hline
SVR                             & \multicolumn{1}{c|}{0.140}          & 0.152              & \multicolumn{1}{c|}{0.340}          & 0.344              & \multicolumn{1}{c|}{0.118}          & 0.129              & \multicolumn{1}{c|}{0.974}          & 0.972              \\ \hline
\end{tabular}
\end{table*}

\section{RESULT \& DISCUSSION}
This section presents the outcomes obtained from the applied machine learning methods for cardiovascular risk estimation and disease identification. Both regression and classification algorithms were assessed using real-world and synthetically generated datasets to ensure comprehensive and reliable conclusions. The detailed analysis for each model is presented below, supplemented with Explainable AI visualizations, residual plots, ROC curves, confusion matrices, and tables.

\subsection{Result Analysis}
Table \ref{tab:cn} summarizes the performance metrics for different classifiers in identifying cardiovascular disease, where Random Forest demonstrated the highest effectiveness. It attained top scores in F1-score (0.977), recall (0.981), precision (0.974), MCC (0.952),  and accuracy (0.976) on the SMOTE data, showcasing its strength in balanced classification tasks. Gradient Boosting and Decision Tree followed closely with 0.970 and 0.974, accuracy scores, respectively. On the other hand, TabNet performed the worst, achieving an accuracy of 0.418 on the artificial dataset.  The TabNet is susceptible to hyperparameter settings and relatively low effectiveness on tiny structured data, which may be the cause of this underperformance, impairing its capacity to generalize. Overall, the outcomes reinforce the superiority of the Random Forest model and demonstrate how class balancing via SMOTE enhances predictive performance.
\vspace{2mm}

Table \ref{tab:rg} outlines the regression model outcomes for heart disease risk estimation. Among all models, Linear Regression yielded the best performance, achieving R2 scores of 0.984 for original data and 0.992 for SMOTE data, as well as the lowest MAE of 0.036 on original and 0.066 on SMOTE data, and MSE values of 0.034 on original and 0.061 on SMOTE data, demonstrating excellent accuracy and minimal prediction error. Models like XGBoost and Random Forest also performed well, with Random Forest upholding consistent R² scores of 0.980 on both datasets. On the other hand, CatBoost produced the least favorable results, having the highest error metrics, MAE of 0.858 and ,MSE of 1.144 and an R2 score of 0.750 on actual data. These results highlight the superior capability of Linear Regression in forecasting heart disease risk.
\vspace{2mm}

\subsection{Confusion Matrix Representations}
\begin{figure} [htbp]
	\centering 
\subfigure [Real]{\label{fig:a}\includegraphics[height = 4.2cm, width=4.34cm]{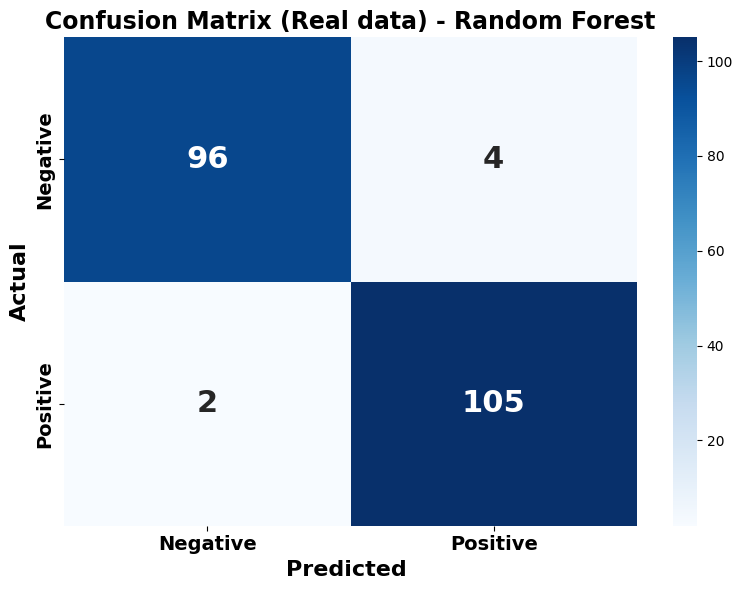}} \hspace{-4mm}
\subfigure [SMOTE]{\label{fig:b}\includegraphics[height = 4.2cm, width=4.34cm]{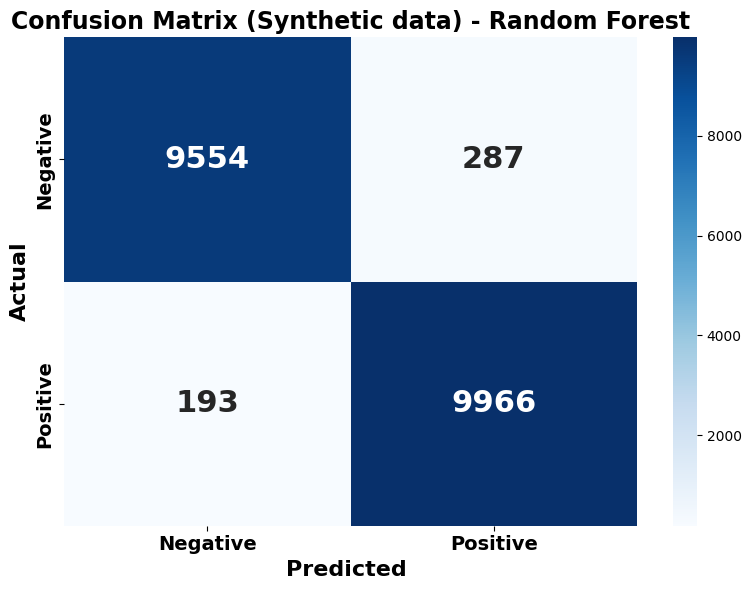}}	
		\caption{Confusion Matrices ((a). Real Data and (b). Synthetic data) of Random Forest for Heart Disease Diagnosis} 
	\label{fig:cm}
\end{figure}

Figure \ref{fig:cm} shows the Random Forest's confusion matrix both before and after SMOTE was applied. In Figure \ref{fig:a}, while the model misclassified two positive samples as negative and four negative samples as positive before SMOTE, it accurately predicted 105 positive and 96 negative samples. In contrast, Figure \ref{fig:b} illustrates improved performance after SMOTE, accurately identifying 9966 samples as positive and 9554 samples as negative. The model incorrectly classified 287 negatives as positives and 193 positives as negatives. The use of SMOTE enhances prediction balance by reducing false negatives, thus improving sensitivity and overall model effectiveness in recognizing positive cases. This reflects strong classification capability with minimal misclassification.

\subsection{ROC Curve Representations}
\begin{figure} [htbp]
	\centering 
	\includegraphics[height=5.2cm, width=8.5cm]{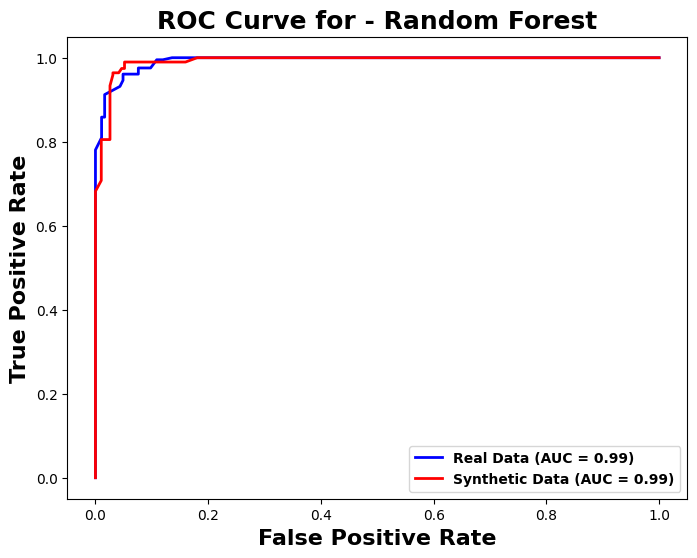}	
		\caption{ROC Curve of Random Forest for Heart Disease Diagnosis on Both Real \& Synthetic Data} 
	\label{figure:rc}
\end{figure}

Figure \ref{figure:rc} shows the Random Forest's ROC curve both before and after SMOTE was applied. The ROC curve displays the False Positive Rate (FPR) versus the True Positive Rate (TPR) across various threshold settings, serving as a tool to evaluate classification performance. The curves for real and synthetic datasets closely align, each achieving an AUC score of 0.99. This elevated AUC suggests how well the model can differentiate across classes. The slight variation between the curves implies that SMOTE effectively handles the issue of class distribution without degrading the model's predictive quality.

\subsection{Residual Analysis Representations}

\begin{figure} [htbp]
	\centering 
\subfigure [Real]{\label{fig:c}\includegraphics[height = 3.8cm, width=4.2cm]{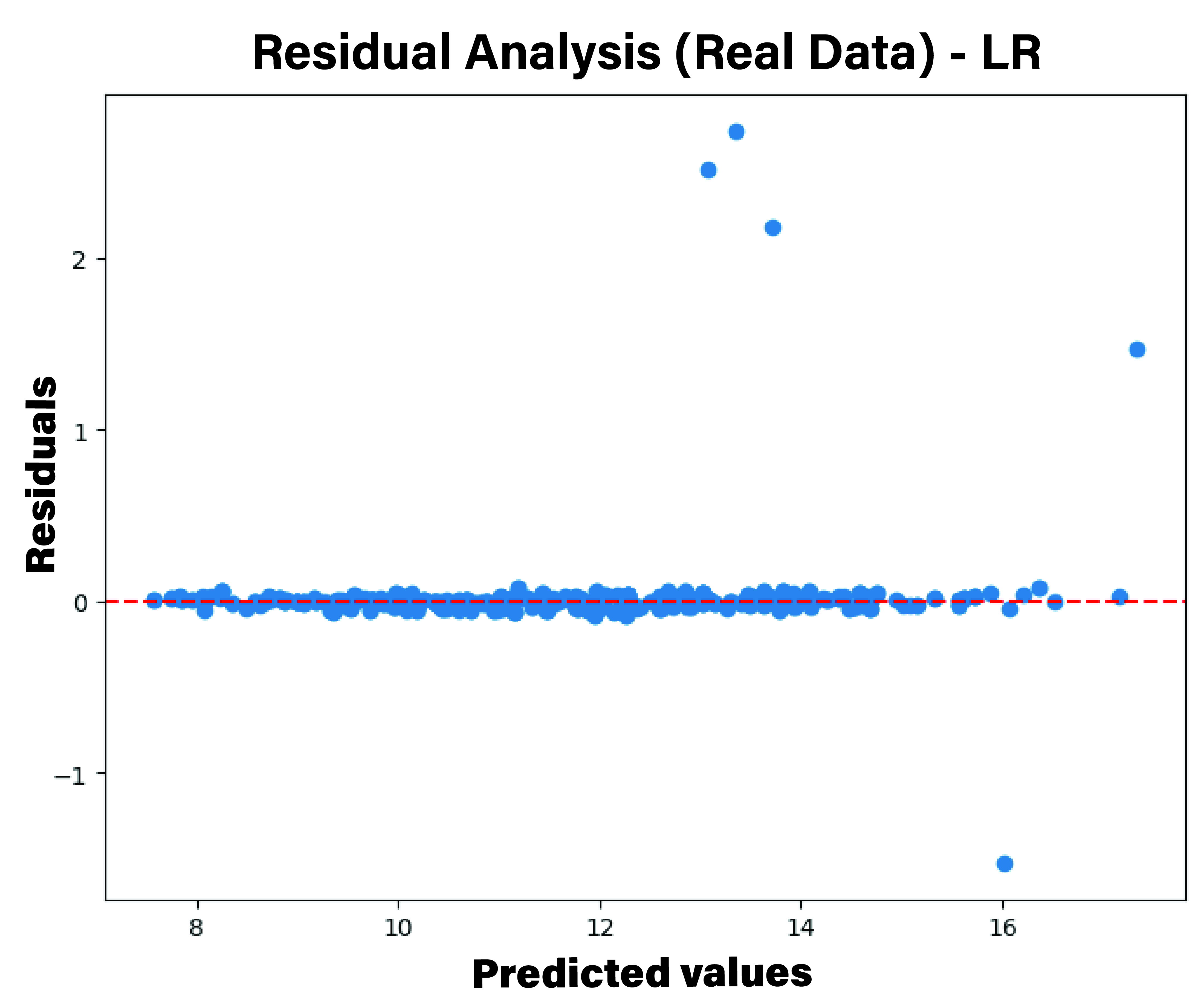}} \hspace{-4mm}
\subfigure [SMOTE]{\label{fig:d}\includegraphics[height = 3.8cm, width=4.2cm]{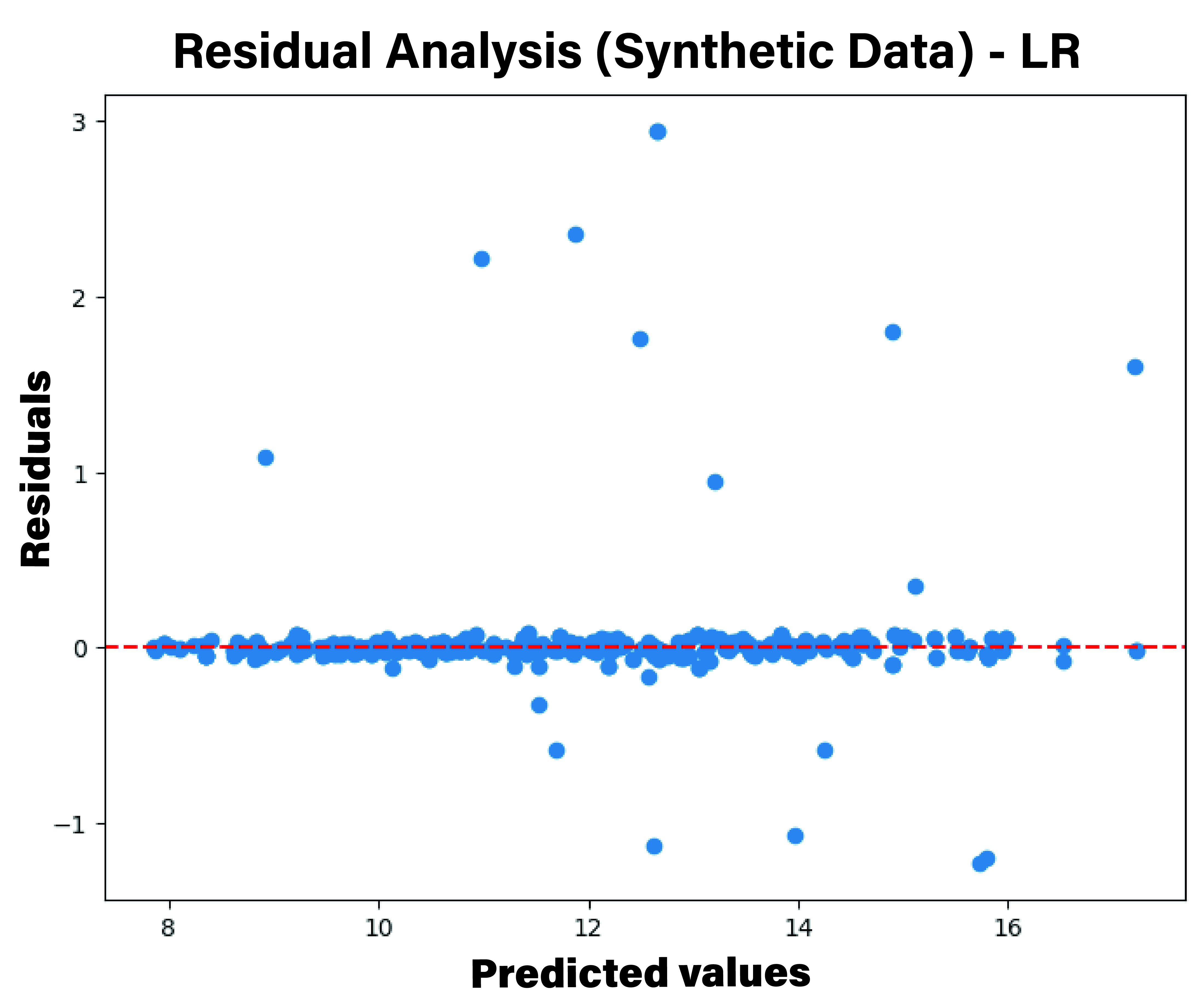}	
}
		\caption{Residual Analysis ((a). Real Data and (b). Synthetic data) of Logistic Regression for Heart Disease Prognosis} 
	\label{fig:ra}
\end{figure}

Figure \ref{fig:ra} presents the Linear Regression model's residual analysis for both Real and SMOTE-generated data. In Figure \ref{fig:c}, residuals—defined as the differences between predicted and actual values, are plotted against predicted outputs. Similarly, Figure \ref{fig:d} displays a scatter plot comparing predicted values with residual differences, where the majority of points lie near zero, signifying minimal prediction bias. The visualizations demonstrate that residuals are mostly concentrated around zero, indicating effective model performance. Nonetheless, a few scattered outliers exist, reflecting instances of higher prediction error.

\subsection{Explainable AI (SHAP and LIME) Representations}

\begin{figure} [htbp]
	\centering 
\fbox{\subfigure [Diagnosis]{\label{fig:e}\includegraphics[height=5.8 cm, width=4cm]{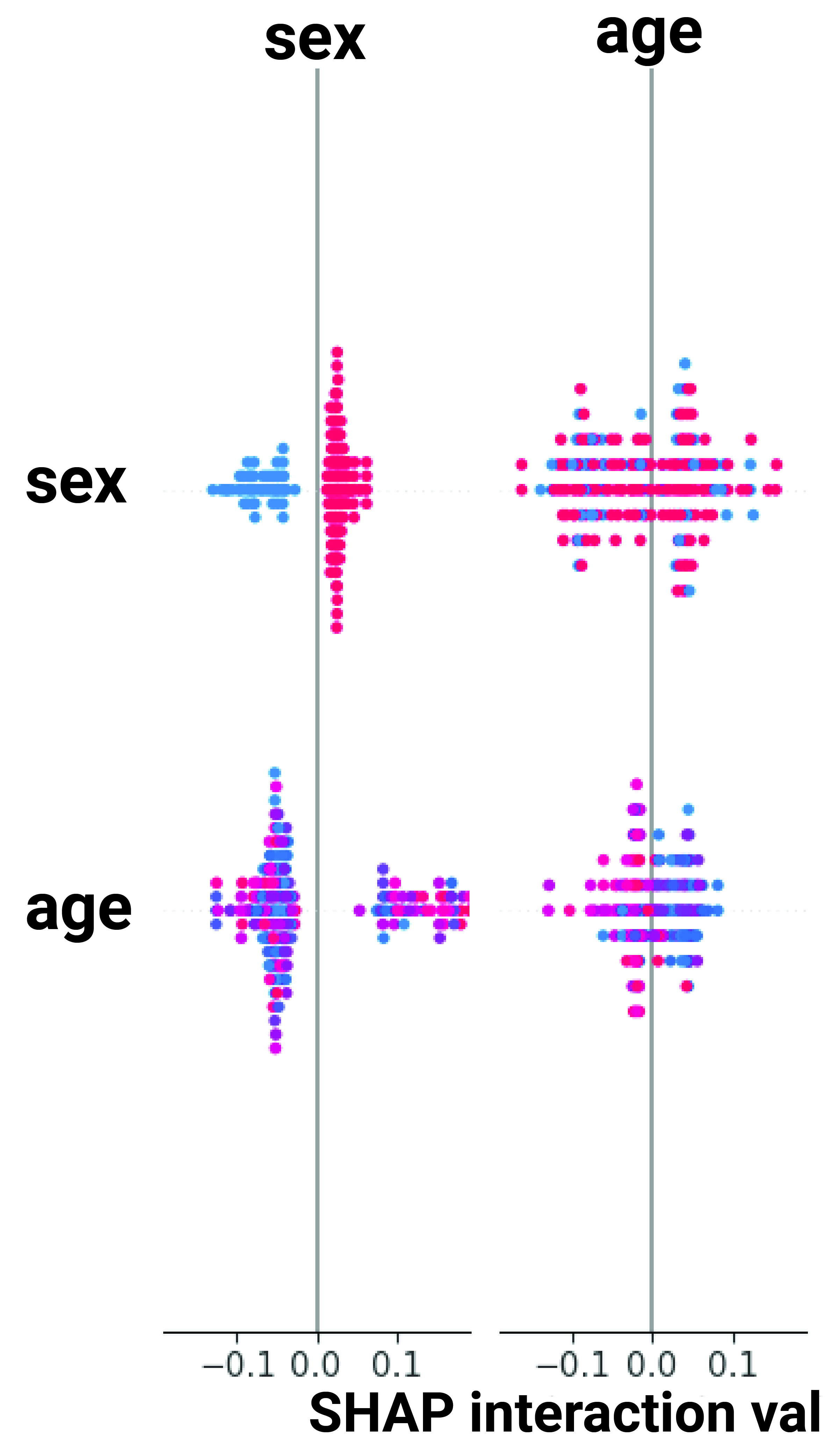}}}
\fbox{\subfigure [Prognosis]{\label{fig:f}\includegraphics[height=5.8 cm, width=4cm]{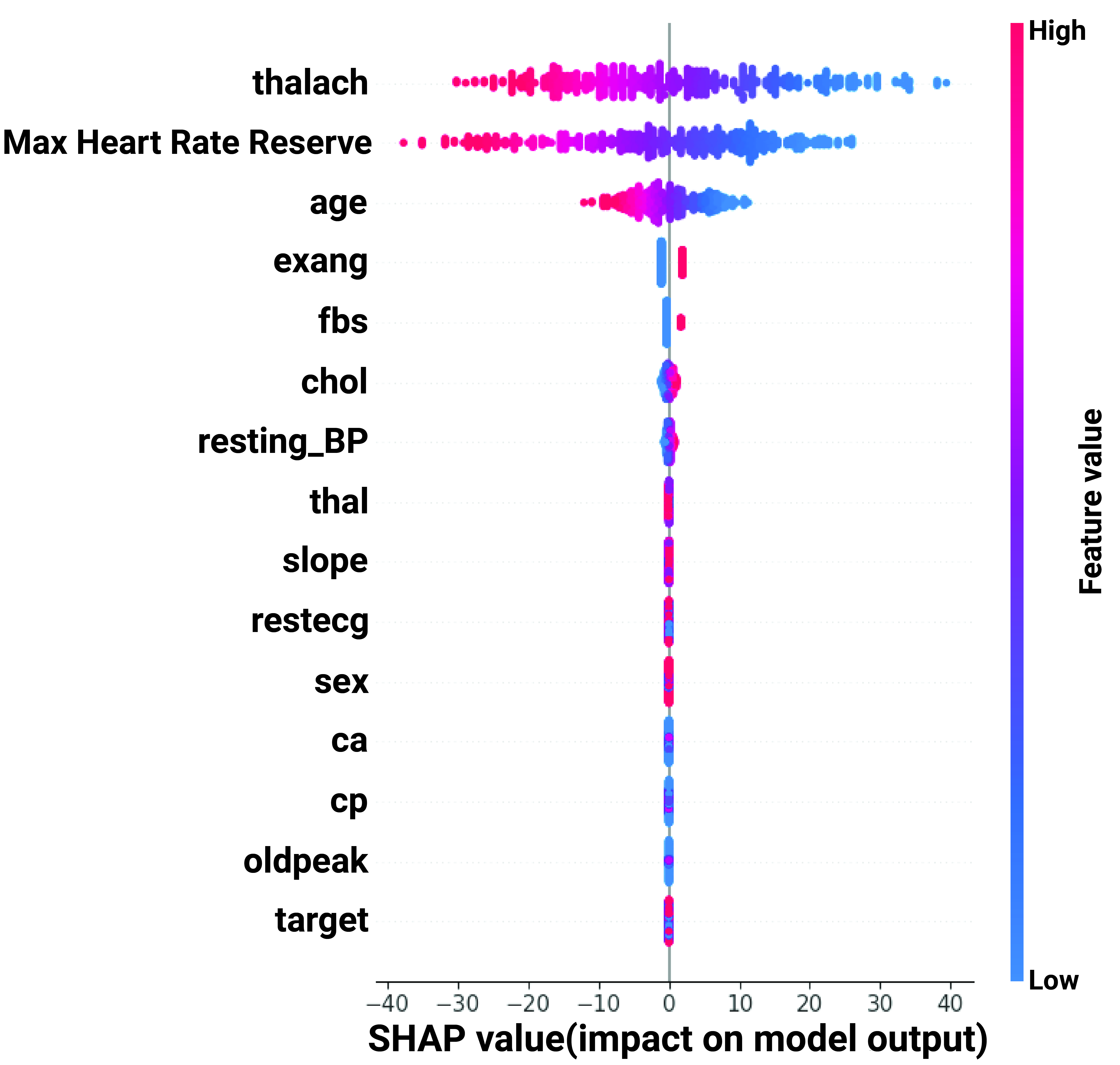}}}	
		\caption{SHAP for Random Forest ((a). Diagnosis) and Logistic Regression ((b). Prognosis) After Balancing the Dataset with SMOTE} 
	\label{fig:sp}
\end{figure}

Figure \ref{fig:e} depicts the SHAP visualization for the RF model after balancing the dataset, showing how significant features like “sex” and “age” influence the model’s forecasts in detecting cardiovascular disease. The SHAP interaction values, which show how each feature affects the model's output, are displayed on the horizontal axis. Positive predictions are pushed toward the heart disease positive class, while negative predictions are pushed toward the negative class. \vspace{1mm}

Figure \ref{fig:f} illustrates the SHAP plot for the Linear Regressor model applied to the regression task after SMOTE.  The figure displays, on the y-axis, the relative relevance of each feature's contribution to the model's predictions. The x-axis displays the SHAP values, indicating both the intensity and direction of feature impact. Features such as age, Max Heart Rate Reserve, and thalach (maximum heart rate) appear as the most significant influential contributors to the regressor's prediction behavior.

\begin{figure} [htbp]
	\centering 
	\fbox{\includegraphics[height=3.73cm, width=8.4cm]{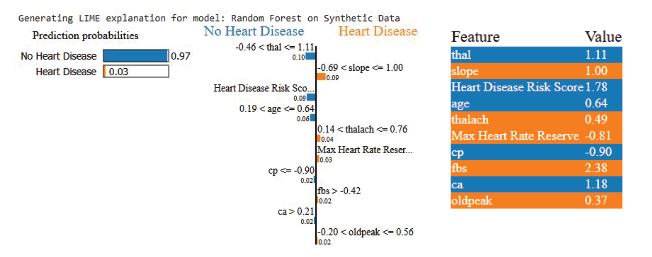}}	
		\caption{LIME for Random Forest with SMOTE for Heart Disease Diagnosis} 
	\label{fig:l1}
\end{figure}

Figure \ref{fig:l1} shows the Random Forest's LIME summary plot following SMOTE classification, indicating the interpretability of the classifier's forecast for the identification of heart disease.  The likelihood of forecasting “Heart Disease" (0.03) and “No Heart Disease" (0.97) for a particular case is displayed.  The prediction's primary determinants, together with their corresponding values and contribution weights, are highlighted in the right panel.  The model's choice is influenced by the values of several key features, including "thal," "slope," and "Heart Disease Risk Score," which can have a favorable or negative effect.

\begin{figure} [htbp]
	\centering 
	\fbox{\includegraphics[height=3.73cm, width=8.4cm]{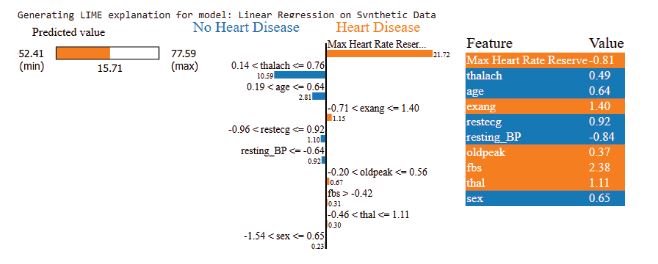}}	
		\caption{LIME for LR with SMOTE for Heart Disease Prognosis} 
	\label{fig:l2}
\end{figure}

Figure \ref{fig:l2} illustrates the LIME of the Linear Regressor model after utilizing SMOTE in the prediction phase. The instance's anticipated value, which is heavily influenced by important characteristics, falls between "No Heart Disease" and "Heart Disease." Features like age, thalach, and  Max Heart Rate Reserve are highlighted in the bar along with their positive and negative contributions, demonstrating their importance in influencing the model's result. To put the forecast in context, the feature values are listed in the accompanying table.

\subsection{Discussion \& Limitation}
Through performance comparison among multiple models and techniques, Random Forest emerged as the most reliable method for cardiovascular disease classification, while Linear Regression proved most effective for risk estimation. The model's strong performance metrics indicate its potential for dependable use in practical applications. We also benchmarked our framework against prior research discussed in the related work section. As reflected in Table \ref{tab:com}, our model surpasses all previously examined models. Nonetheless, a notable limitation of the proposed framework lies in its reliance on the underlying dataset's quality and variability. Even though SMOTE was utilized to counteract data imbalance, the data may still lack the diversity found in actual clinical conditions, which could restrict the model’s capacity to generalize effectively.

\begin{table}[htbp]
\setlength{\tabcolsep}{3.2pt}
\renewcommand{\arraystretch}{2}
\centering
\caption{Comparison with State-of-the-Art}
\label{tab:com}
\begin{tabular}{|c|c|c|}
\hline
\textbf{Ref.} & \multicolumn{1}{c|}{\textbf{Model}}                                                           & \multicolumn{1}{c|}{\textbf{Findings}} \\ \hline
\cite{r16}        & \begin{tabular}[c]{@{}c@{}}Adaptive Boosting (AdaBoost)\end{tabular}                           & 95\% Accuracy                           \\ \hline
\cite{r10}        & Deep Neural Network                                                                            & 95.30\% Accuracy                              \\ \hline

\cite{r12}       & Neural Network                                                                                              & 92\% AccuracyAccuracy                     \\ \hline
\cite{r15}        & Decision Tree                                                                                & 95.76\% Accuracy                        \\ \hline

\cite{r14}      & Multilayer Perceptron                                                                                 & 87.28\% Accuracy                           \\ \hline

\begin{tabular}[c]{@{}c@{}}\textbf{Proposed} \\ \textbf{Model}\end{tabular}                   & \begin{tabular}[c]{@{}c@{}}\textbf{Random Forest for Diagnosis}\\ \textbf{Logistic Regression for Prognosis} \end{tabular}                                           & \begin{tabular}[c]{@{}c@{}}\textbf{97.6\% with RF and} \\ \textbf{99.2\% R² with LR}\end{tabular}                                                                                                                \\ \hline
\end{tabular}
\end{table}

\section{CONCLUSION and FUTURE WORK}

This research introduces a dependable and interpretable machine learning approach for identifying and forecasting Heart conditions, achieving an accuracy of 0.976 using Random Forest for detection and a 0.992 R² score via Linear Regression for risk prediction. By integrating powerful algorithms with interpretability tools like SHAP and LIME, the approach ensures high accuracy while shedding light on the most influential risk indicators. SMOTE was instrumental in addressing class imbalance, leading to enhanced results on the synthetic dataset and reinforcing the model's practical relevance. This work emphasizes machine learning’s role in early detection and prognosis, and the value of explainable AI for trustworthy medical decision-making. Looking ahead, future work may involve enriching the dataset with additional clinical features, conducting long-term studies for progression modeling, and evaluating the model's effectiveness across different demographic and healthcare settings. \vspace{3mm}

\bibliographystyle{ieeetr}
\bibliography{ref.bib}

@misc{dataset,
  author={ S.N. Mahsa },
  title        = {Heart Disease Dataset},
  year         = {2020},
  publisher    = {Kaggle},
  howpublished = {Kaggle, Available at: {https://www.kaggle.com/datasets/snmahsa/heart-disease}}
}

@article{r1,
  title={Optimized ensemble learning approach with explainable AI for improved heart disease prediction},
  author={ Mienye, Ibomoiye Domor and others },
  journal={Information},
  volume={15},
  number={7},
  pages={394},
  year={2024},
  publisher={MDPI}
}

@article{r10,
  title={Predictive analytics of congestive heart failure using deep neural networks: a comparative study},
  author={ Singh, Amit Kumar and others },
  journal={Computers, Materials \& Continua},
  volume={68},
  number={2},
  pages={2165--2181},
  year={2021}
}

@article{r12,
  title={Harnessing AI for early detection of cardiovascular diseases: Insights from predictive models using patient data},
  author={Husnain, Ali and others},
  journal={International Journal for Multidisciplinary Research},
  volume={6},
  number={5},
  year={2024}
}

@inproceedings{hossen2025july, 
author = {Hossen, Md Sabbir and others}, 
title = {Social Media Sentiments Analysis on the July Revolution in Bangladesh: A Hybrid Transformer Based Machine Learning Approach}, booktitle = {Proceedings of the 17th International Conference on Electronics, Computers and Artificial Intelligence (ECAI)}, year = {2025}, 
publisher = {IEEE} }

@article{r14,
  title={Effective heart disease prediction using machine learning techniques},
  author={Bhatt, Chintan M and others},
  journal={Algorithms},
  volume={16},
  number={2},
  pages={88},
  year={2023},
  publisher={MDPI}
}

@article{r15,
  title={A comprehensive study of machine learning for predicting cardiovascular disease using Weka and SPSS tools},
  author={Abuhaija, Belal and others},
  journal={International Journal of Electrical and Computer Engineering},
  volume={13},
  number={2},
  pages={1891},
  year={2023},
  publisher={IAES Institute of Advanced Engineering and Science}
}

@article{r16,
  title={A Technical Comparative Heart Disease Prediction Framework Using Boosting Ensemble Techniques},
  author={Nissa, Najmu and others},
  journal={Computation},
  volume={12},
  number={1},
  pages={15},
  year={2024},
  publisher={MDPI}
}

@article{chandola2008work,
  title={Work stress and coronary heart disease: what are the mechanisms?},
  author={Chandola, Tarani and others},
  journal={European heart journal},
  volume={29},
  number={5},
  pages={640--648},
  year={2008},
  publisher={Oxford University Press}
}

@article{hoffman2013global,
  title={The global burden of congenital heart disease},
  author={Hoffman, Julien IE},
  journal={Cardiovascular journal of Africa},
  volume={24},
  number={4},
  pages={141--145},
  year={2013},
  publisher={Clinics Cardive Publishing}
}

@article{mondesir2016diabetes,
  title={Diabetes, diabetes severity, and coronary heart disease risk equivalence: REasons for Geographic and Racial Differences in Stroke (REGARDS)},
  author={Mondesir, Favel L and others},
  journal={American heart journal},
  volume={181},
  pages={43--51},
  year={2016},
  publisher={Elsevier}
}

@article{mocumbi2011challenges,
  title={Challenges on the management of congenital heart disease in developing countries},
  author={Mocumbi, Ana Olga and others},
  journal={International journal of cardiology},
  volume={148},
  number={3},
  pages={285--288},
  year={2011},
  publisher={Elsevier}
}

@article{chang2022artificial,
  title={An artificial intelligence model for heart disease detection using machine learning algorithms},
  author={Chang, Victor and others},
  journal={Healthcare Analytics},
  volume={2},
  pages={100016},
  year={2022},
  publisher={Elsevier}
}

@misc{tribune,
  author    = {Tribune Desk},
  title     = {Experts: One in four adults in Bangladesh suffer from hypertension},
  howpublished = {Dhaka Tribune},
  note      = {Available at: https://www.dhakatribune.com/bangladesh/health/360195/experts-one-in-four-adults-in-bangladesh-suffer},
  year      = {2024}
}

@article{chandrasekhar2023enhancing,
  title={Enhancing heart disease prediction accuracy through machine learning techniques and optimization},
  author={Chandrasekhar, Nadikatla and Peddakrishna, Samineni},
  journal={Processes},
  volume={11},
  number={4},
  pages={1210},
  year={2023},
  publisher={MDPI}
}

@inproceedings{strawberry,
  author       = {M. Sabbir Hossen and others},
  title        = {A Hybrid Machine Learning Approach Utilizing CNN Feature Extraction with Traditional Classifier to Identify Strawberry Leaf Diseases},
  booktitle    = {4th International Conference on Electrical, Computer and Communication Engineering (ECCE)},
  organization = {IEEE},
  year         = {2025}
}

@article{dwivedi2023explainable,
  title={Explainable AI (XAI): Core ideas, techniques, and solutions},
  author={Dwivedi, Rudresh and others},
  journal={ACM Computing Surveys},
  volume={55},
  number={9},
  pages={1--33},
  year={2023},
  publisher={ACM New York, NY}
}

@article{pp3,
  title={An extensive experimental analysis for heart disease prediction using artificial intelligence techniques},
  author={Rohan, D and others},
  journal={Scientific Reports},
  volume={15},
  number={1},
  pages={6132},
  year={2025},
  publisher={Nature Publishing Group UK London}
}

@article{pp2,
  title={Ensemble learning with explainable AI for improved heart disease prediction based on multiple datasets},
  author={Ganie, Shahid Mohammad and others},
  journal={Scientific reports},
  volume={15},
  number={1},
  pages={13912},
  year={2025},
  publisher={Nature Publishing Group UK London}
}

@article{pp1,
  title={Performance evaluation of optimal ensemble learning approaches with PCA and LDA-based feature extraction for heart disease prediction},
  author={Rabbi, Md Sakhawat Hossain and others},
  journal={Biomedical Signal Processing and Control},
  volume={101},
  pages={107138},
  year={2025},
  publisher={Elsevier}
}

@misc{hossen2025brain,
      title={An Efficient Deep Learning Framework for Brain Stroke Diagnosis Using Computed Tomography (CT) Images}, 
      author={Md. Sabbir Hossen and others},
      year={2025},
      eprint={2507.03558},
      note  = {arXiv, cs.CV, Available at: {https://arxiv.org/abs/2507.03558}
}}

\end{document}